\documentclass[runningheads]{llncs}
\usepackage{graphicx}
\usepackage{hyperref}
\usepackage{amsfonts,amssymb}
\hypersetup{colorlinks,
	linkcolor=red,%
	citecolor=blue}
\begin{document}
\title{Frame Aggregation and Multi-Modal Fusion Framework for Video-Based Person Recognition}
\titlerunning{Frame Aggregation and Multi-Modal Fusion}
\author{Fangtao Li \and Wenzhe Wang \and Zihe Liu \\ \and Haoran Wang \and Chenghao Yan \and Bin Wu{*}}
\institute{Beijing University of Posts and Telecommucations \\ \email{\{lift, wangwenzhe, ziheliu, wanghaoran, \\ chenghao\_yan, wubin\}@bupt.edu.cn}}

\authorrunning{F. Li et al.}
\maketitle              
\begin{abstract}

Video-based person recognition is challenging due to persons being blocked and blurred, and the variation of shooting angle. Previous research always focused on person recognition on still images, ignoring similarity and continuity between video frames. To tackle the challenges above, we propose a novel Frame Aggregation and Multi-Modal Fusion (FAMF) framework for video-based person recognition, which aggregates face features and incorporates them with multi-modal information to identify persons in videos. For frame aggregation, we propose a novel trainable layer based on NetVLAD (named AttentionVLAD), which takes arbitrary number of features as input and computes a fixed-length aggregated feature based on the feature quality. We show that introducing an attention mechanism into NetVLAD effectively decreases the impact of low-quality frames. For the multi-model information of videos, we propose a Multi-Layer Multi-Modal Attention (MLMA) module to learn the correlation of multi-modality by adaptively updating correlation Gram matrix. Experimental results on iQIYI-VID-2019 dataset show that our framework outperforms other state-of-the-art methods.\footnote{This work is supported by the National Key Research and Development Program of China (2018YFC0831500), the National Natural Science Foundation of China under Grant No.61972047, the NSFC-General Technology Basic Research Joint Funds under Grant U1936220 and the Fundamental Research Funds for the Central Universities (2019XD-D01).}

\end{abstract}
\section{Introduction}

Person recognition in videos is a basic task for video understanding. Different from person recognition in images, video-based person recognition has the following difficulties and challenges: (1) The expressions, angles, and clarity of the person may vary in videos. (2) Videos are more informative than images, which means the model that can use information effectively tends to be more complicated. 
Most of the current research is still stuck in recognition based on frames. In general, when extended to video-based person recognition, the essence of the study is to use frames as units of recognition, but with the addition of pooling or voting to aggregate the results. However, this aggregation method neglects the quality of frames, which may lead to misidentification. 

To tackle this problem, a natural idea is to remove low-quality frames during data preprocessing. However this method places high requirements on how to measure the quality of frames. It will also reduce the generalization of the model. Another widely used method is to equip models with the ability to learn frame quality. To this end, NeXtVLAD~\cite{lin2018nextvlad} apply a non-linear activation function as a simple attention mechanism, and GhostVLAD~\cite{zhong2018ghostvlad} implicitly achieves this goal by "Ghost" clusters, which can help the network ignore low-resolution or low-quality frames. However, these heuristic rules are incapable of the complex cases in the real world. 

In some extreme cases, the resolution of the face in the video is so low that the clear face cannot be recognized, so we can only rely on other information such as audio, text, \textit{etc.}, to help identify of the person. However, most methods integrate multi-modal information through pooling, concatenation, or heuristic rules, which is over-simplified in complex real-world scenarios.

In this work, we propose a Frame Aggregation and Multi-Modal Fusion (FAMF) framework for video-based person recognition. To measure the quality of different frames, we propose a Vector of Locally Aggregated Descriptors with cluster attention (AttentionVLAD) algorithm. In order to enhance the robustness of the model and better handle the extreme case where the face is completely obscured, we introduce a Multi-Layer Multi-Modal Attention (MLMA) module to use multi-modal information for joint projections. Finally, we evaluate the performance of our framework through experiments on the iQIYI-VID-2019 dataset~\cite{liu2019iqiyi}. The experimental results show the effectiveness of our framework.

In short, our contribution is as follows:
\begin{itemize}
	\item [1.]
	We propose a FAMF framework to recognize persons in videos, which considers both frame quality and multi-modal information for end-to-end recognition.
	\item [2.]
	For frame aggregation, we propose a new trainable feature aggregation module named AttentionVLAD, which uses attention mechanism to adjust the weight of aggregated output. For multi-modal fusion embedding, we propose a MLMA module inspired by attention mechanism, which uses continuous convolutional layers to obtain the multimodal attention weights. 
	\item[3.]
	We verify the effectiveness of the proposed framework on the iQIYI-VID-2019 dataset. Our approach outperforms other state-of-the-art methods without any model begging or data augmentation.
	
\end{itemize}

\section{Related Work}

\subsubsection{Person Recognition}
Person recognition has been widely studied in different research areas. In general, face recognition is the most important sub-task for person recognition, while person re-identification (person Re-ID), speaker identification, \textit{etc.}, can also be regarded as part of person recognition. 

As one of the most studied sub-questions of person recognition, face recognition has achieved great success along with the arising of deep
learning. For face recognition on still images, many algorithms perform quite well. ArcFace~\cite{deng2019arcface} reached a precision of 99.83\% on LFW~\cite{huang2008labeled} dataset, which outperforms the human performance. 
Person re-identification aims to identify the pedestrian in case the face is blurred or invisible. Many algorithms~\cite{song2018region}\cite{zheng2019joint} performs well on this task. However, most of these algorithms use the body as a crucial feature for identification, and they have not addressed the problem of changing clothes yet. 
For speaker recognition, i-vector~\cite{dehak2010front} is a widely used algorithm for a long time. More recently, d-vector based on deep learning~\cite{li2017deep} became increasingly popular and performs well. However, the large amount of noise in the real scene along with the misalignment of the speakers and the characters in the video is the restricted factors that affect the further application of this technology.

\subsubsection{Multi-Modal Fusion}
\setlength{\parskip}{-0em} 
For multi-modal fusion, early works concatenate embeddings
to learn a larger multi-modal embedding. However this may lead
to a potential loss of information between different modalities. Recent studies on learning multi-modal fusion embeddings apply neural network to incorporate modalities. Tensor Fusion Network (TFN)~\cite{zadeh2017tensor} calculates the outer-product of video, audio and text features to represent comprehensive features. Liu \textit{et al.}~\cite{liu2018efficient} developed a low rank method for building tensor networks to reduce computational complexity caused by outer-product. Adversarial representation learning and graph fusion network~\cite{DBLP:conf/aaai/Mai0X20} are also applied for multi-modal fusion. In recent years, some researchers apply attention mechanism on multi-modal fusion. Liu \textit{et al.}~\cite{liu2018iqiyi} proposed a Multi-Modal Attention (MMA) module, which reweight modalities through a Gram Metrix of self-attention. Tsai \textit{et al.}~\cite{tsai2019multimodal} built a multi-modal transformer to learn interactions between modalities. Xuan \textit{et al.}~\cite{DBLP:conf/aaai/XuanZCYY20} proposed a cross-modal attention network, which concentrates on specific locations, time segments and media adaptively. 
\subsubsection{Feature Aggregation}
In the early stages of the development of image recognition,  many encoding methods were proposed to aggregate image descriptors into a template representation. Along with the era of deep learning, some researchers integrate these algorithms into neural networks, such as NetFV~\cite{tang2016deep} and NetVLAD~\cite{arandjelovic2016netvlad}. Recently, Lin \textit{et al.}~\cite{lin2018nextvlad} proposed a NeXtVLAD model for video classification, which decreases the number of parameters by group convolution. Zhong \textit{et al.}~\cite{zhong2018ghostvlad} introduced a feature aggregation module named GhostVLAD for set-based face recognition, which dramatically downweights blurry and low-resolution images by ghost clusters, and GhostVLAD surpassed the state-of-the-art methods on IJB-B~\cite{whitelam2017iarpa} dataset. 
\begin{figure}[tb]
	\setlength {\abovedisplayskip}{0cm}
	\setlength {\belowcaptionskip}{-0cm}
	\includegraphics[width=5in]{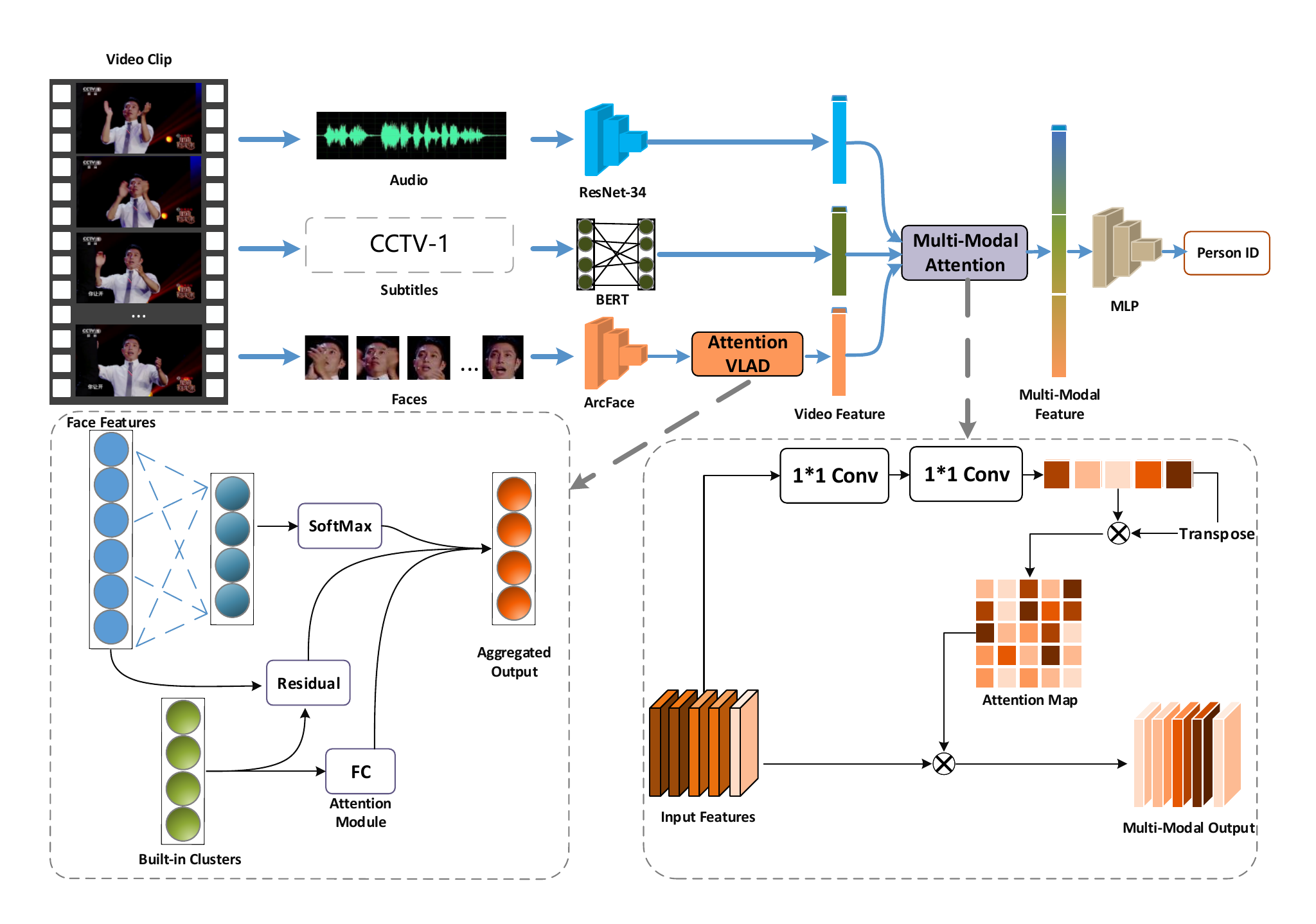}
	\centering
	\caption{An overview of the process of identifying the person in a given video clip. The FAMF framework consists of a feature extractor, a frame feature aggregation module, a multi-modal fusion module and a classifier. Face embeddings are aggregated by AttentionVLAD and fused with other modal embeddings by MLMA.} \label{framework}
\end{figure}
\section{Our Framework}

\subsection{Overview}
In this paper, we regard person recognition as a classification task, which means given a piece of video, we expect our model can determine the label of the main characters in the video. As shown in Fig. \ref{framework}, our framework consists of four modules:

\subsubsection{Feature Extractor}
In this stage, we mainly perform video pre-processing and feature extraction. Multi-modal information such as face, audio, body, text, \textit{etc.}, will be extracted from the input video. Then we use pre-trained models to extract embedding respectively. 

\subsubsection{Frame Aggregation Module}
A video is composed of continuously changing frames, which means if a face is detected in a certain frame, it is likely that its adjacent frames also contain the face with the same label, despite their angle, expression or clarity may be varied. To measure the quality of faces and eliminate the complexity caused by the different number of faces in different clips, we propose a frame feature aggregation module based on NetVLAD. Taking any number of features as input, this module can generate a fixed-length embedding as a video-level face feature. More details can be found in Sec.~\ref{AttentionVLAD}.

\subsubsection{Multi-modal Fusion Module}
Multi-modal features of a video clip are likely to be complementary and redundant. In this stage, we apply a multi-modal attention module to learn the weight of corresponding multi-modal features and readjust the multi-modal features according to the weight. This module will be introduced in detail in Sec.~\ref{MLMA}.
\subsubsection{Classifier}
Based on the premise that we focus on supervised person recognition in video, we use a Multi-Layer Perceptron (MLP) as a classifier to predict the identity of the person.

\subsection{AttentionVLAD for frame aggregation} \label{AttentionVLAD}
One key problem of video-based person recognition is that people do not appear in all frames of the video, the clarity and the quality of each frame is also varied. If all detected face features are concatenated as input, although all the information of visual cues are retained, the model will be memory-cost, which has a great negative impact on model training and storage. In previous research, NetVLAD is a widely used layer to aggregate low-level features. In this section, we propose an improved frame features aggregation layer, AttentionVLAD, to improve the impact of low-quality frames.
\subsubsection{NetVLAD}
Given $N$ $D$-dimensional features $X \in \mathbb{R}^{N*D}$ extracted from the input video clip, which are likely to be extracted in an arbitrary manner using an arbitrary pre-trained model, such as $N$ frame features extracted by ResNet~\cite{he2016deep}, or $N$ face features extracted by ArcFace~\cite{deng2019arcface}, all these $N$ features are aggregated into feature $V \in \mathbb{R}^{D*K}$ where $K$ is a hyper-parameter indicating the number of trainable clusters. The formula is shown as below:
\begin{equation}
V(j,k) = \sum_{i=1}^{N} \alpha_{k}(x_i) (x_{i}(j) - c_{k}(j)),
\end{equation}
\begin{equation}
\alpha_{k}(x_i) = \frac{e^{a_{k}^{T} x_{i} + b_{k}}}{\sum_{k^{'}=1}^{K} e^{a_{k}^{T} x_{i} + b_{k}}},
\end{equation}
where $\{x_{i} \in \mathbb{R}^{D}, i = 1...N\}$ is the input features, $\{c_k \in \mathbb{R}^{D}, i = 1...K\}$ is the $N$-dimensional centre of cluster $k$, $\{a_k\}$, $\{b_k\}$ and $\{c_k\}$ are trainable parameters. $\alpha_{k}(x_{i})$ is the soft-assignment weight of input feature $x_i$ for cluster $c_k$.

\subsubsection{AttentionVLAD}

Based on NetVLAD, we proposed a novel feature aggregation module with cluster attention named AttentionVLAD, which is applied to  adjust the weight of low-quality feature in the aggregation stage. In our experiment, AttentionVLAD is deployed on face modal, since face is the crucial modal for person recognition.
Given the number of trainable clusters $K$ and $N$ $D$-dimensional features $X \in \mathbb{R}^{N*D}$, which are the output of the feature extractor, where $N$ refers to the number of faces, respectively. The aggregated output feature $V \in \mathbb{R}^{D*K}$ can be generated as below:

\begin{equation}
V(j,k) = \sum_{i=1}^{N} \phi(c_{k}) \alpha_{k}(x_i) (x_{i}(j) - c_{k}(j)),
\end{equation}
\begin{equation}
\alpha_{k}(x_i) = \frac{e^{a_{k}^{\top} x_{i} + b_{k}}}{\sum_{k^{'}=1}^{K} e^{a_{k}^{\top} x_{i} + b_{k}}},
\end{equation}
where $x_i$ refers to the $i$-th feature, ${c_k}$ is the $N$-dimensional anchor point of cluster $k$, $\{a_k\}$, $\{b_k\}$ and $\{c_k\}$ are trainable parameters. $\alpha$ is the soft-assignment weight of input feature for cluster c. $\phi(c_{k})$ is an attention weight for cluster $k$ implemented by a full-connected layer. $\alpha_{k}(x_{i})$ is the soft-assignment weight of input feature $x_i$ for cluster $c_k$, as in NetVLAD~\cite{arandjelovic2016netvlad}. Note that we calculate the similarity by the cluster center, and adjust the weight of aggregated feature, which is different with self-attention.

The improvement of AttentionVLAD lies in the cluster attention function, which can learn the feature quality implicitly. To learn the impact of face quality on prediction result, an easy way is to measure the weight at the instance level, \textit{e.g.} adding a non-linear activation for each feature. However, due to the complexity of the feature, the activation function may not learn the weight as expected. Another way is to use trainable clusters to adjust weights implicitly. This idea first appeared in GhostVLAD~\cite{zhong2018ghostvlad}, which uses $G$ ghost clusters to collect features of low-quality and drop them in the output. In fact, GhostVLAD~\cite{zhong2018ghostvlad} can be regarded as a specialization of AttentionVLAD when $K$ of the cluster attention weight $\phi(c_k) $ in proposed AttentionVLAD are 1 and $G$ of them are 0. In other words, compared to the binary cluster weight of 0 or 1 in GhostVLAD~\cite{zhong2018ghostvlad}, AttentionVLAD can achieve better results with adaptive stepwise cluster weights.

Another explanation of AttentionVLAD is that it introduces the information of trainable clusters themselves into the output. A key point of NetVLAD~\cite{arandjelovic2016netvlad} is that it uses residuals of the input and the trainable clusters to represent the aggregation template. Although residuals can retain the input information well, the distribution and scale of input is completely ignored. which infer the quality of features. In order to rehabilitate this information in aggregation output, we use cluster weight as a scalar to fuse the residuals between inputs and clusters, which is calculated by the attention mechanism.

\subsection{MLMA for Multi-modal Fusion} \label{MLMA}
Given a video clip with a main character appearing, in the ideal case, we can only rely on the detected face to determine the identity of the character. However, the face of person will inevitably be blurred or blocked, or we can not even detect a clear face in the entire video. In this situation, we can use other information about the video to predict the identity. To eliminate the redundancy of multi-modal information and use complementary information as much as possible, we devise a Multi-Layer Multi-Modal Attention (MLMA) module to fuse the multi-modal features according to the weights.

Given features $X \in \mathbb{R}^{(K_{1}+K_{2})*D}$, where $K_1$ refers to the number of the clusters in AttentionVLAD, and $K_2$ refers to the number of the other multi-modal feature of the source video, such as audio, body, text, textit{etc.}, and $D$ is the dimension of features. The weight of each modal can be yielded through the continuous convolution layer and a soft-max activation:

\begin{equation}
Y_i = \sum_{j}^{K_{1}+K_2}X_{j}\frac{e^{Z_{j,i}}}{\sum_{i}{e^{Z_{j,i}}}},
\end{equation} 
\begin{equation}
Z=(W_{F_1}W_{F_2}X)^{\top} (W_{F_1}W_{F_2}X),
\end{equation}
where $Z$ is the Gram matrix of multi-modal attention weight, $W$ are trainable parameters. $Y_i$ represents the summation of products between cross-correlation weights and feature $X_{j}$.

The MLMA is inspired by SAGAN~\cite{zhang2019self} and MMA~\cite{liu2018iqiyi}. For multi-modal feature in videos, MLMA can efficiently capture the inter-modal correlation. From the perspective of enhancing correlation and reducing redundancy, a modal of feature that is inconsistent with other modal of feature will be amplified after fusion by MLMA, implying that they are weakly correlated. In another point of view, if a certain modal  is missing, MLMA will try to replace it by another strongly related and similarly distributed feature, which is important to compensate the loss when some features are missing.

Our motivation is to improve the matrix acquisition method of Gram matrix of multi-modal features. MMA~\cite{liu2018iqiyi} has shown that using Gram matrix can better incoporate multi-modal features, since it is simpler and more effective in capturing the feature correlation. However, it simply uses a convolution layer to reduce the number of feature channels at one-time, and then multiply with the transpose of embedding to get Gram matrix. Residual connection and L2-regularization are also applied. We have found in experiments that if features are processed by two consecutive convolutional layers, better results can be achieved, and the effect of residual connection and L2-regularization term is not obvious. Therefore, we did not use these tricks.

\section{Experiments}
\subsection{Dataset}
\begin{figure}[tb]
	\setlength {\abovedisplayskip}{0cm}
	\setlength {\belowcaptionskip}{-0cm}
	\includegraphics[width=4.8in]{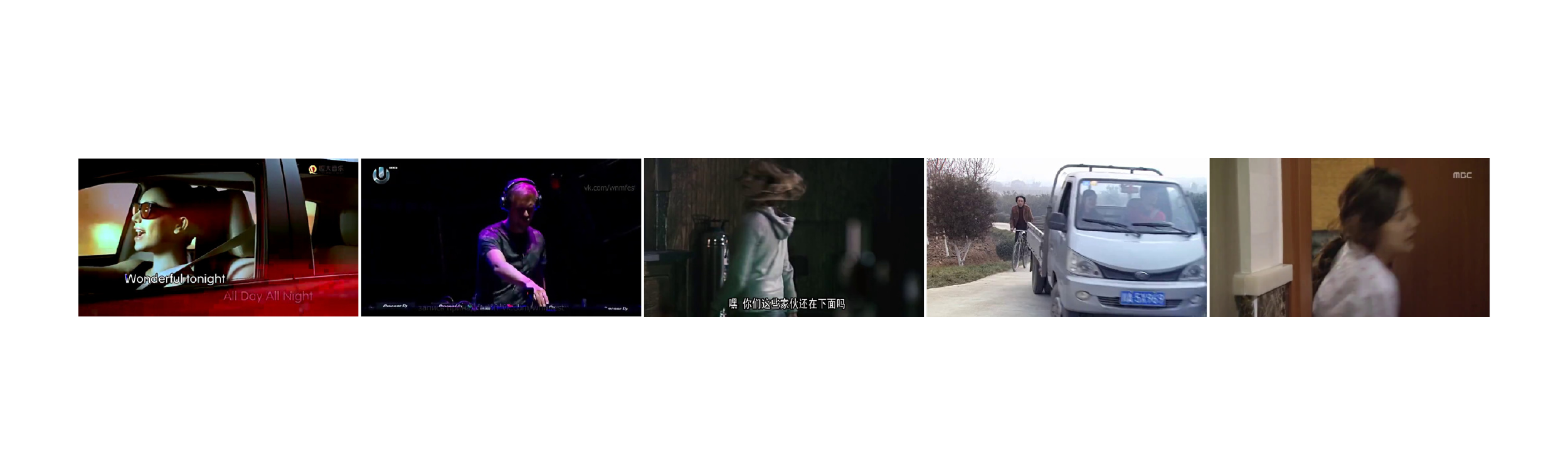}
	\centering
	\setlength{\abovecaptionskip}{0pt}
	\setlength{\belowcaptionskip}{0pt}
	\caption{Some challenging cases of iQIYI-VID-2019 dataset. From left to right: occlusion, weak light, invisible face, small face, and blur.} \label{dataset}
\end{figure}

The iQIYI-VID-2019 dataset~\cite{liu2019iqiyi} is a large-scale benchmark for multi-modal person identification. It contains more than 200K videos of 10,034 celebrities. To the best of our knowledge, it is the largest video dataset for person recognition so far. In 2019, the whole dataset including test set was released in ACM MM workshop. The dataset contains both video clips and the official features of face, head, body and audio. We removed the noise data from the training set in the preprocessing stage, and do not apply any data augmentation in experiment. All models are trained on the training set and evaluated on the validation set. Some challenging cases are shown in Fig. \ref{dataset}.

\subsection{Results}
We use Mean Average Precision (mAP) to evaluate the performance of models, mAP can be calculated as :
\begin{equation}
mAP(Q) = \frac{1}{|Q|}\sum_{i=1}^{|Q|}\frac{1}{m_i}\sum_{j=1}^{n_i}Precision(R_{i,j}),
\end{equation}
where $Q$ is the set of person ID to retrieve, $m_i$ is the number of positive examples for the $i$-th ID, $n_i$ is the number of positive examples within the top $k$ retrieval results for the $i$-th ID, and ${R_{i,j}}$ is the set of ranked retrieval results from the top until getting $j$ positive examples. In our implementation, only the top 100 retrievals are kept for each person ID.

\begin{table}[tb]
	\caption{Comparison with state-of-the-art methods on iQIYI-VID-2019 dataset. A higher mAP value is better.}\label{tab1}
	\begin{center}
		\begin{tabular}{c|c|c|c}
			\hline
			Method&Frame Aggregation & Multi-Modal Fusion & mAP \\
			\hline
			Liu \textit{et,al.}~\cite{liu2018iqiyi}&NetVLAD &MMA &0.8246\\
			GhostVLAD~\cite{zhong2018ghostvlad}&GhostVLAD &Concat &0.8109\\
			NeXtVLAD~\cite{lin2018nextvlad}&NeXtVLAD &Concat &0.8283\\
			\hline
			
			FAMF-mf&NetVLAD &MLMA &0.8295\\
			FAMF-fa&AttentionVLAD &MMA &0.8610\\
			FAMF(Ours)&AttentionVLAD &MLMA &\textbf{0.8824}\\
			
			\hline
		\end{tabular}
	\end{center}
\end{table}
Table \ref{tab1} shows the performance comparison of our model with the baseline models on iQIYI-VID-2019 dataset~\cite{liu2019iqiyi}. We mainly compared with some state-of-the-art person recognition methods. NetVLAD+MMA means we use NetVLAD~\cite{arandjelovic2016netvlad} for temporal aggregation and Multi-Modal Attention~\cite{liu2018iqiyi} for multi-modal feature fusion. This is also the official baseline model of iQIYI-VID-2019~\cite{liu2019iqiyi}. GhostVLAD and NeXtVLAD are implemented with reference to ~\cite{zhong2018ghostvlad} \cite{lin2018nextvlad}, respectively.  FAMF-mf means we only apply AttentionVLAD for frame aggregation, while FAMF-fa means we replace MMA~\cite{liu2018iqiyi} with MLMA. For a complete version, the result is shown as in \textit{row} 6, which adopts AttentionVLAD for frame aggregation and MLMA for multi-modal fusion.

\begin{figure}[tb]
	\setlength {\abovedisplayskip}{0cm}
	\setlength {\belowcaptionskip}{-0.5cm}
	\includegraphics[width=5in]{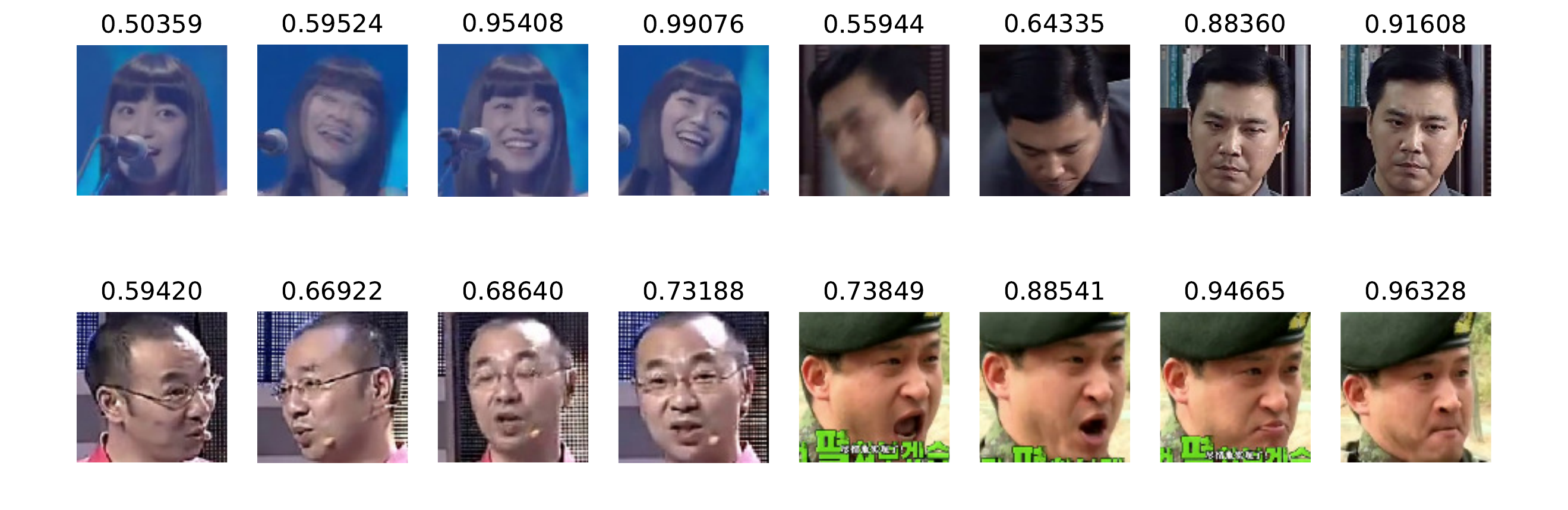}
	\centering
	\caption{The weight of different faces in a same video is shown on the above of faces. The contribution of blurred or blocked faces is significantly reduced, and the weight of clear, unobstructed faces is still close to 1.} \label{weight}
\end{figure}

We select some video clips randomly in the test set, and print the weight of different keyframes, as shown in Fig. \ref{weight}. \textit{row} 1 shows two videos in which faces that are obscured or blurred (\textit{column} 1, 2, 5, 6) have significantly lower weights compared to clear and unobscured faces. Two videos in \textit{row} 2 show that the frames of low resolution have significant low weight (\textit{column} 1 \textit{to} 4). These examples indicating the cluster attention mechanism has the ability to learn image quality.

\setlength{\parskip}{-1mm}
\subsection{Implementation Details}
\paragraph{Preprocessing}
During data preprocessing, for visual features of each video, such as face, body, \textit{etc.}, if the number of features is greater than the frame number $N$ (we set $N$ = 24), then we sampled $N$ features for each video randomly without repeating, otherwise we sampled repeatedly until the number of features is equal to $N$. Note that the number $N$ can be different during training and testing stage. For audio information, we use the official audio embedding directly. For text or subtitle of a certain video clip, we choose one frame and use the return value of Baidu OCR API as the text, then take the output of pre-trained BERT~\cite{devlin2018bert} as text feature. If a certain modal is missing, we use a zero vector as the corresponding embedding. 
\setlength{\parskip}{0mm}
\paragraph{Model Settings}
The classifier consists of 3 fully-connected(FC) layers, of which the first two layers are used as input and hidden layers, and the output of the last layer is mapped to the labels. Between two adjacent FC layers, we add a Batch-Normalization layer. Dropout is not applied. The dimension of all hidden layers is set to 4096, while the output dimension of the last FC layer is 10,034. The hyper-parameter $K$ of AttentionVLAD is 8, and in MLMA, the dimension of two hidden layers are 128 and 32, respectively.

\paragraph{Training Strategy}
All of the models are optimized by Adam Optimizer, and the loss function is Cross-Entropy Loss. The learning rate of AttentionVLAD and the other parameters are initialized as 0.04 and 0.004, respectively. After training for 50 epochs, the learning rates are divided by 10 for every 10 epochs. At training stage, the size of mini-batch is 4,096.

\begin{table}[tb]
	\caption{Experimental results for using different modality information.}\label{tab2}

	\begin{center}
		\begin{tabular}{l|c}
			\hline
			Cues &  mAP\\
			\hline
			Face &  0.8656 \\
			Face+Body &  0.8776 \\
			Face+Audio & 0.8688 \\
			Face+Text & 0.8685 \\
			Face+Audio+Text & 0.8385 \\
			Face+Audio+Body & \textbf{0.8824}\\
			Face+Body+Text & 0.8669 \\
			Face+Audio+Body+Text &0.8761\\
			\hline
		\end{tabular}
	\end{center}
	
\end{table}
\subsection{Ablation Study}

\paragraph{AttentionVLAD}
Firstly we analyze the performance of different aggregation methods. From Table \ref{tab1}, we can find that AttentionVLAD outperforms the other aggregation modules such as NetVLAD~\cite{arandjelovic2016netvlad} (\textit{row} 4 \textit{vs. row} 5). As mentioned in Sec~\ref{AttentionVLAD}, AttentionVLAD can drop low-quality features in steps. Under the premise of using only face features, the cluster attention mechanism can improve the representation.
 
\paragraph{MLMA}
For multi-modal information fusion, we mainly compared results with MMA~\cite{liu2018iqiyi}. The improvement of the effect can be considered as being affected by the two-layer convolutional layer. It can extract more information than the one-layer convolution in the original method. In the case that the increased memory overhead can be ignored, it can improve the mAP by 0.02 to 0.8824 (\textit{row} 4 \textit{vs. row} 6).

\begin{figure}[tb]
	\setlength {\abovedisplayskip}{0cm}
	\setlength {\belowcaptionskip}{-0.5cm}
	\includegraphics[width=1.6in]{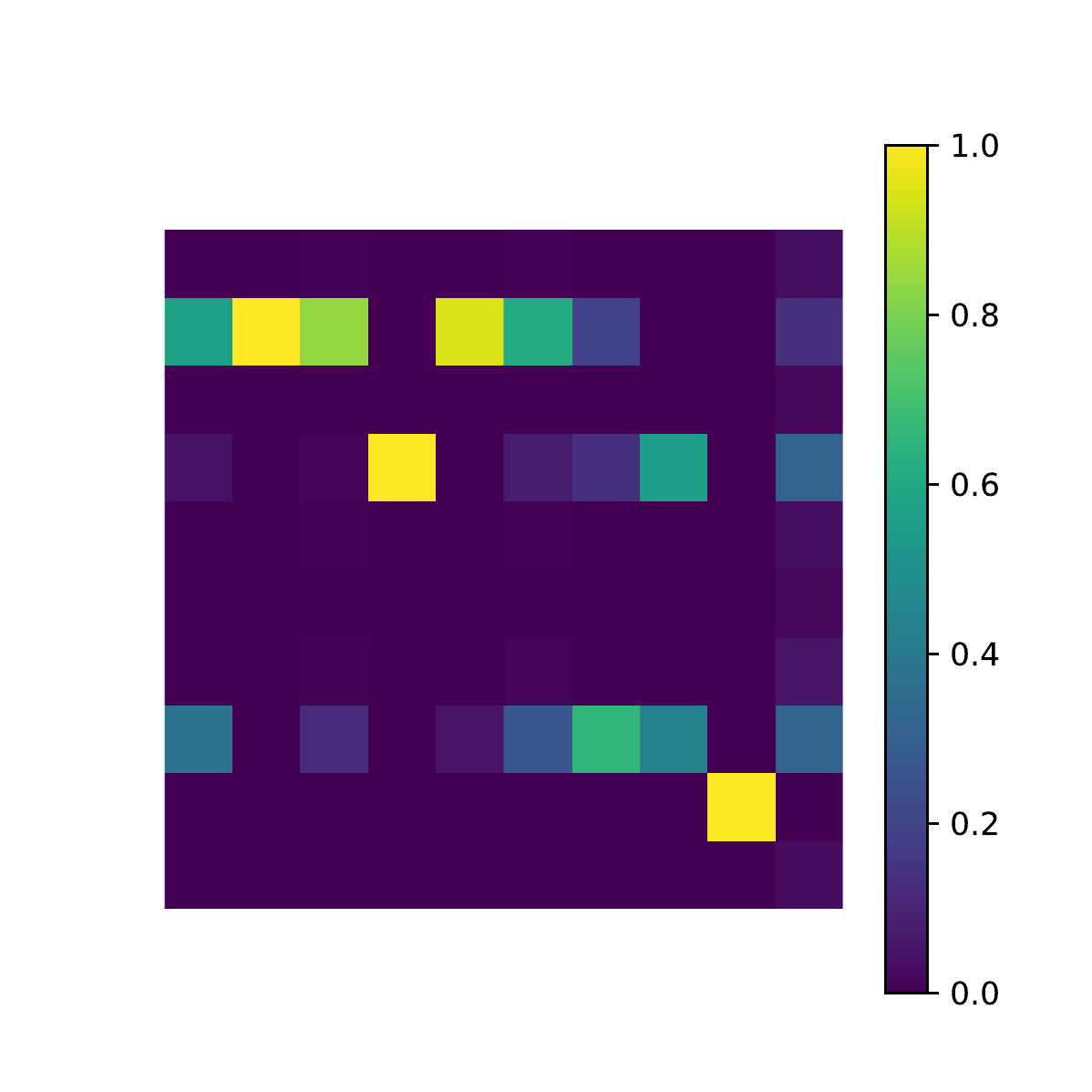}
	\centering
	\caption{Attention weight of multi-modal features. The first eight columns, the ninth column, and the tenth column of the matrix show the cross-correlation weights of face, audio, body, respectively.} \label{fig4}
\end{figure}

\paragraph{Multi-modal Information}
The effect of multi-modal information is shown in Table \ref{tab2}. We assume that face feature is of most importance in person identification, while other features are relatively auxiliary. Taking face feature as a baseline, adding audio feature can raise the performance significantly. We presume it is because audio feature is not strongly correlated with the face feature, which adds less redundant information to enhance the robustness of the model. The attention weight is shown as Fig. \ref{fig4}, where the \textit{row} 1 to \textit{row} 8 refer to the 8 features from AttentionVLAD, the last row refers to the audio feature, and the last row refers to body feature, respectively. The result shows that body feature tends to be replaced by face feature, while the audio feature has a high probability to be independent to concatenate with other features.
\paragraph{Hyper Parameters}
Finally, we discuss the impact of the number of clusters $K$ on the performance. We compare the results for $K$ = 2, 4, 8, 16, respectively. Results show that when $K$ is 2 or 4, the model performs not well (\textit{mAP} = 0.7952 \textit{vs.} 0.8824), proving that the aggregate feature does not effectively filter out the information in each frame feature. The model achieves the best performance when $K$ = 8. In theory, further increasing $K$ can still improve the effect. However, it may cause the problem of out of memory. When $K$ = 16 (\textit{mAP} = 0.8323), in order to train the model, we have to reduce the batch size.
\setlength{\parskip}{-3mm}
\section{Conclusion}
In this paper, we first propose a novel FAMF framework for video-based person recognition. A new feature aggregation module named AttentionVLAD, and a multi-modal feature fusion module MLMA are then proposed. Given a video clip, we first extract multi-modal feature by pre-train model, then for face feature, AttentionVLAD is applied to generate a fix-length template. The AttentionVLAD uses cluster attention mechanism to learn face quality. Then a MLMA module is applied to fuse multi-modal features, which reweights multi-modal features by two convolution layers. The experiment results on iQIYI-VID-2019 dataset show that our model outperforms other person recognition methods. 

\bibliographystyle{splncs04}
\bibliography{ref.bib}

\end{document}